\documentclass[final]{article}

 \PassOptionsToPackage{numbers, compress}{natbib}
\usepackage{nips_2016}

\usepackage[utf8]{inputenc} % allow utf-8 input
\usepackage[T1]{fontenc}    % use 8-bit T1 fonts
\usepackage{hyperref}       % hyperlinks
\usepackage{url}            % simple URL typesetting
\usepackage{booktabs}       % professional-quality tables
\usepackage{amsfonts}       % blackboard math symbols
\usepackage{amsfonts,amsmath,amsbsy,bm,color,verbatim,graphicx,subfig,bibentry}
\usepackage{hyperref}
\usepackage{algorithm}
\usepackage[noend]{algpseudocode}

\input{macros.sty}
\graphicspath{{Figures/}}

\title{Sparse Regression and Adaptive Feature Generation for the Discovery of Dynamical Systems}

\author{
 Chinmay S. Kulkarni \texttt{chinmayk@mit.edu} \\
 Department of Mechanical Engineering, MIT \\ %Massachusetts Instiute of Technology \\
 6.860: Statistical Learning Theory and Applications, Fall 2018, MIT
}

\begin{document}
\maketitle

\begin{abstract}
We study the performance of sparse regression methods and propose new techniques to distill the governing equations of dynamical systems from data.
%In this work, we study the theoretical and experimental performance of some well-established sparse regression methods, and propose two new techniques to achieve the overarching goal of distilling the governing equations of dynamical systems from data.
We first look at the generic methodology of learning interpretable equation forms from data, proposed by \citet{brunton2016discovering}, followed by performance of LASSO for this purpose. We then propose a new algorithm that uses the dual of LASSO optimization for higher accuracy and stability.
%e first look at the generic methodology of learning interpretable equation forms from data, first proposed by \citet{brunton2016discovering}. The first half of this work deals with the case of fixed feature space. We study the theoretical aspects and the potential pitfalls of the most popular sparse regression algorithm: LASSO; applied to the problem of model discovery.
%To alleviate the shortcomings of LASSO, we propose a new methodology which leverages the uniqueness of the solution of the dual problem of the LASSO regression.
%Through this methodology, we are able to select appropriate equation terms even from a set of highly correlated possible options.
In the second part, we propose a novel algorithm that learns the candidate function library in a completely data-driven manner to distill the governing equations of the dynamical system.
%In the second part of this manuscript, we propose a novel algorithm that adapts the candidate feature space with data to optimally capture the dominant components of the dynamical system, starting with an empty feature space.
This is achieved via sequentially thresholded ridge regression (STRidge \cite{rudy2017data}) over a orthogonal polynomial space.
%This algorithm does not impose sparsity on the feature space, but allows for optimal feature selection.
The performance of the three discussed methods is illustrated by looking the Lorenz 63 system and the quadratic Lorenz system.%, two of the most popular chaotic dynamical systems.
%Finally, the performance of the three discussed methods (LASSO, dual LASSO and adaptive feature space growth with STRidge) are illustrated by looking the Lorenz 63 system, which is one of the most popular multi-dimensional chaotic dynamical system.
\end{abstract}

\section{Introduction and Overview} \label{sec:intro}

The role of data today is no longer limited to the verification of first-principles-derived-models but it is also being used to learn such models. This is particularly important in non-autonomous nonlinear dynamical systems that describe a multitude of problems from science and engineering.
%There has been a research thrust in this direction in the last few decades \citep{crutchfield1987equations,schmidt2009distilling,bongard2007automated}.
Recent methods leverage the fact that most dynamical equations governing physical systems contain a few terms, making them sparse in high-dimensional nonlinear function space \citep{brunton2016discovering,rudy2017data}. By constructing an appropriate feature library based on the data coordinates, one can apply sparse regression to discover the governing equations of the dynamical system. Initial work in this field has mainly focused on the behavior of the approach with regards to noise, multi-fidelity data etc., however few have tried to improve upon the sparse regression algorithm at the core of the approach.

This is exactly the first focus area of the current work. We first look at the sparse regression method most commonly employed in this field: LASSO \cite{tibshirani1996regression}. We point out the theoretical underpinnings of LASSO along with accuracy bounds.
Although LASSO works well for uncorrelated features, for highly correlated it tends to choose a feature at random from each of the correlated groups \cite{gauraha2017dual}. 
%Although LASSO works well with assured fast convergence rates for (almost) uncorrelated features, it converges slower for highly correlated features, and tends to choose a feature at random from each of the correlated groups \cite{gauraha2017dual}.
% The accuracy dependence on the hyperparameter is also amplified in this case.
%Note that when will build a feature library from the data coordinates, the higher order features are more and more correlated with each other, where LASSO may potentially run into difficulties.
When the feature library is built from data, higher order features are more correlated with each other, where LASSO potentially runs into difficulties.
To alleviate these difficulties, we propose to solve the dual of LASSO to learn the governing equations.
% To alleviate the potential difficulties that LASSO might face, we propose to solve the dual of the LASSO optimization to learn the governing equations.
Even in the case of correlated features, the dual LASSO has a unique solution, which then allows us to correctly choose the present features.
%In order to obtain accurate model fitting, we compute an $L_2$ regularized regressor only on the present features.
The second part of this work deals with the case when the exact function blocks that describe the dynamical system are not present in the feature library. We propose and investigate an approach to handle such cases by using an appropriate family of orthogonal functional basis to span the feature library.
As the result may not be sparse in terms of the orthogonal functions, we resort to sequentially thresholded least squares (STRidge) algorithm \cite{rudy2017data} that does not impose sparsity, but chooses the dominant features. 
%This allows us to to add new components to the feature space easily while discarding those that do not have any projection of the dynamical system along them. This ensures that the size of the feature space does not grow quickly. However, a major implication of transforming the basis to an orthogonal one is that the dynamical system may not be sparse in these basis. Thus instead of using sparse regression techniques, we resort to sequential thresholding least squares (STRidge) algorithm \cite{rudy2017data} that does not necessitate sparsity, but chooses the most dominant features. We use this algorithm iteratively, while adding or removing appropriate features to dynamically adapt our feature space for the best approximation of the equations from data.

%These approaches are demonstrated on one of the most popular chaotic nonlinear systems in atmospheric and fluid flows: the Lorenz 63 system \cite{lorenz1963deterministic}.
These approaches are demonstrated on the Lorenz 63 system \cite{lorenz1963deterministic}and the quadratic Lorenz system \cite{eminauga2015modified}.
We compare the performance of the LASSO and dual LASSO, and then use the approach of data-driven orthogonal regression to learn the feature library. %Finally, we conclude with summary and a discussion over the potential directions to build upon the current work.

\subsection{General Methodology}\label{sec:general_method}

% We now briefly describe the general procedure behind our method for model discovery.
Let us assume that we have $n$ state space parameters ($x_1, \ldots, x_n$), with measurements for $x_i$ and $\dot{x_i} = dx/dt$ at times $t = 1, \ldots, T$ (denoted by a superscript).
If only state observations are available, the rate parameters can be computed using finite difference.
% In some cases, the rate parameters may be directly available (e.g. one might measure position and velocity both).
This is followed by constructing a non-linear library of features using the state space parameters. The span of these features now describes the feature space.
%The method to construct this feature space is a modeling choice.
Typically we would construct this feature space through a class of functions that are dense in the space that our dynamical system lives in.
In this work we assume a polynomial feature library. %, however the methodology is agnostic towards the choice of functional basis and would apply to any other feature library as well.
%In this work, we use the class of polynomials.
% Note that as we start including higher order polynomials, the successive features are more and more correlated.
After constructing the feature library (say $X$), we formulate the regression problem as $\dot{X} = XW + \varepsilon$, where $\dot{X}_{(t, j)} = \dot{x_j}^t$ and $W$ are the unknown weights, with $\varepsilon$ being the noise.

\begin{figure}[h]
	\centering
	 \includegraphics[width=\linewidth]{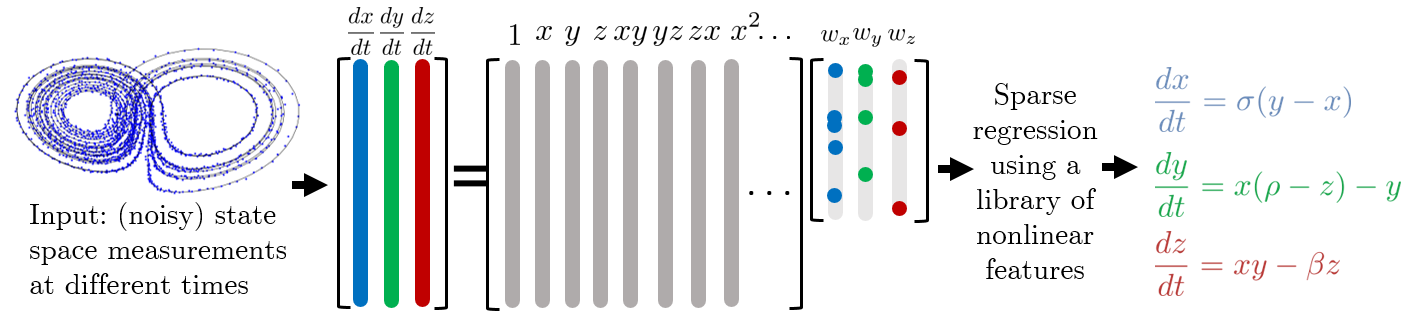}
	\caption{Schematic of the generic methodology used.}
	\label{fig:schematic}
\end{figure}

Often, dynamical models have a functional form where the terms live in a smaller manifold in the feature space.
%In other words, not all the features in the library that we consider are required to explain the dynamical model.
We thus use sparse regression to pick the relevant features.
%Once the relevant features are obtained through sparse regression, we perform component-wise thresholding to further ensure that the features with extremely small coefficients are not considered.
These features, with their corresponding coefficients describe the functional form of the governing equations. This procedure is pictorially illustrated in figure \ref{fig:schematic}. For noisy observations, we carry out data filtering and smoothing at the input step and perform multiple sparse regression solves until convergence.
% In such cases, successive solves may be performed until the sparsity of the feature space remains unchanged.
However we do not focus on this, and we assume that we have noise free observations.

\vspace*{-0.25in}

%A similar procedure works even for PDEs. In this case, we write the system as $\partial u / \partial t = \mathcal{L}(u)$. The operator $\mathcal{L}$ contains the spatial gradients (such as $\partial u / \partial x, \partial u / \partial y$ etc.). A similar system is formulated as above. However, we need to take care as we now have an entire spatial field as a state variable at a particular time. Hence, we create a 1D augmented state space description by flattening out this spatial field. The final sparse regression problem can be formulated by constructing a system where the description at a particular time is itself a vector (instead of a scalar, as was the previous case), but with shared parameters. Beyond this, the sparse regression methodology follows through in this new system as well.

\section{Sparse regression over fixed feature space} \label{sec:fixed}

In this case, we assume that the feature library is fixed, and that we wish to find either the exact sparse equation form from this library or the closest approximation to the governing equation only from the terms in the library.
The highest polynomial degree in the feature space ($X$) is $p$. Then, the feature space contains terms of the form $(x^t_1)^{p_1} \ldots (x^t_i)^{p_i} \ldots (x^t_n)^{p_n}$, such that $p_1 + \ldots + p_n \leq p$. The number of solutions to this system (\ie the number of terms in the feature library) is $m = {n+p \choose n} = \frac{(n+p)!}{n!p!}$.
%This implies that $X \in \RR^{T \times m}$.
Whereas, empirically the number of distinct terms in the governing equations is $\mathcal{O}(n)$. Thus even for small enough $p$, the terms in the feature library are much more in number than those to be chosen, which justifies sparse regression to select the features. Let us denote the coefficient matrix, obtained from the sparse regression solve by $W$. The optimization problem with some form of penalty ($\mathcal{P}$) is:
\begin{equation} \lb{reg}
\min_{W} \mathcal{L}(W) = \left[ \( \dot{X} - XW \)^2 + \mathcal{P}(W) \right] \ \ \text{where} \ \ \dot{X} \in \RR^{T \times n} \ , \ X \in \RR^{T \times m} \text{ and, } W \in \RR^{m \times n}
\end{equation}
To further select the features appropriately, we use our knowledge of the underlying physics of the dynamical system. Often, we have knowledge about the general scales of the state space parameters (for example, in the Lorenz system, the state space parameters represent the rate of convection, horizontal temperature variation and the vertical temperature variation). Thus, we know the the typical scales of these parameters from physics. Let us denote the scale of $x_i$ by $L_i$, then $(x^t_1)^{p_1} \ldots (x^t_n)^{p_n} \approx L_1^{p_1} \ldots L_n^{p_n}$. Thus, instead of selecting features based on their coefficient values, we select features by looking at their net magnitude, by replacing the state space parameters with their respective scales.
%This allows us to retain the terms that have the maximum impact on the evolution of the system, instead of only considering terms with the largest coefficients.
We refer to this as `scale based thresholding'.

\subsection{Sparsity through $L_1$ regularization (LASSO)} \label{sec:lasso}

We now look at the use of LASSO for imposing sparsity. As is well-known, the LASSO penalty is $\mathcal{P}(W) = \lambda ||W||_1$, which serves as a convex counterpart to the non-convex $L_0$ norm. \citet{kakade2009complexity} showed that the Rademacher averages for linear regressors with $L_1$ penalty are bounded by $\mathcal{R}_n(\mathcal{F}_W) \leq X W_{\max} \sqrt{\frac{2 \log(m)}{n}}$, where $W_{\max} = \max ||W||_1$. We can then use the contraction property, assuming the Lipschitz constant of the LASSO problem to be $L$ (more details and methods to compute it may be found in \cite{konevcny2016mini,konecny2013semi}) to obtain the bound of Rademacher averages of the loss function class $\mathcal{G}$. Using the symmetrization lemma, this yields a bound on the expected maximum error (uniform deviation), as given by \eref{symm}. 
\begin{equation} \lb{symm}
\E \max_{g \in \mathcal{G}}\left[ \E g(x_i) - \frac{1}{n} \sum_{i = 1}^{n} g(x_i) \right] \leq 2 L XW_{\max} \sqrt{\frac{2 \log(m)}{n}}
\end{equation}
Note that even though we cannot explicitly compute the right hand side, we have $m >>> n$, which renders this accuracy bound impractical. Thus, LASSO in the case of exceedingly high number of features does not provide good accuracy bounds. To alleviate this problem, we first use the SAFE bounds provided by \citet{ghaoui2010safe} to remove the irrelevant features and then apply LASSO over the remaining library. Even after weeding out the absent features, another important task that remains is to choose the hyperparameter $\lambda$. Literature widely suggests the order of $\lambda$ to be $\lambda \approx \sqrt{T\log(m)}$ \cite{bickel2009simultaneous}, however choosing this parameter independent of the degree of correlation is not recommended \cite{dalalyan2017prediction}. It is further shown that the larger the correlation amongst the features, the smaller the value of $\lambda$ \cite{hebiri2013correlations}. \citet{hebiri2013correlations} also prove that LASSO can achieve fast rate bounds (\ie error bounded by $\lambda^2$) even for correlated features. Even though reassuring, in practice this requires significant amount of tuning and cross validation, which is a major issue in the case at hand, as typically we only have sparse and limited observations of the physical states. Further, even though the LASSO fit is unique in case of high correlations, they are problematic for parameter estimation and feature selection \cite{osborne2000lasso,tibshirani2013lasso}. The pitfalls of LASSO (even after removing the irrelevant features using the SAFE rules) are that it requires significant hyperparameter tuning and it is extremely sensitive to $\lambda$ for correlated features (observed empirically). These motivate us to instead formulate a new approach to solve the sparse regression problem.

% Write about choosing / tuning hyperparameter value, and rates of convergence for large polynomial degree.
% Maybe we can first do a LASSO pass, with super low lambda to remove the high order polynomials and then get fast rates of convergence as the remaining polynomials are not super highly correlated.

% Although the other paper claims lasso works well even for highly correlated features, it needs a lot of tuning.

%Problems with LASSO: no analytical solution, problem with correlated variables (which is the case for us), non-unique solution as the problem is not strictly convex (chooses any one of the features from the correlated group).

%\textbf{LASSO works well with correlations, but correlations are problematic for parameter estimation and variable selection, which is what is of interest here. There is always a unique fitted value $XW$, but not a unique $W$.}

\subsection{Sparsity through dual LASSO regression} \label{sec:dual}

To overcome the difficulties in the application of LASSO (along with the SAFE rules), we formulate and solve its dual problem. In order for the LASSO solution (that is, the feature and parameter choices) to be unique, the feature matrix must satisfy the irrepresentability condition (IC) of \citet{zhao2006model} and beta-min condition of \citet{buhlmann2011statistics}. The feature library violates the IC for highly correlated columns, leading to an unstable feature selection. However, even for highly correlated features, the corresponding dual LASSO solution is always unique \cite{tibshirani2013lasso}. The dual problem is given by \eref{dual} \cite{osborne2000lasso}, which is strictly convex in $\theta$ (implying a unique solution).
\begin{equation}\lb{dual}
\max_{\theta} \mathcal{D}(\theta) = ||\dot{X}||_2^2 - ||\theta - \dot{X}||_2^2 \quad \quad \text{such that} \quad \quad ||X^T \theta||_{\infty} \leq \lambda
\end{equation}
Now, let $\hat{W}$ be a solution of \eref{reg} with LASSO penalty and $\hat{\theta}$ be the unique solution to the corresponding dual problem \eref{dual}. Then stationarity condition implies $\hat{\theta} = \dot{X} - X\hat{W} \lb{sta}$.
%%
%\begin{equation}
%%&X^T  \( \dot{X} - X\hat{W} \) + \lambda \text{sign}(\hat{W}) = 0 \lb{kkt1} \\
%\hat{\theta} = \dot{X} - X\hat{W} \lb{sta}
%\end{equation}
Even though LASSO does not have a unique $\hat{W}$, the fitted value $X\hat{W}$ is unique, as the optimization problem \eref{reg} is strongly convex in $XW$ for $\mathcal{P}(W) = \lambda ||W||_1$. We make use of this by first computing a solution to the primal LASSO problem and then computing the unique dual solution by using the primal fitted value and \eref{sta}. Once we have the unique dual solution $\hat{\theta}$, we do feature selection by using the dual active set, which is same as the primal active set with high probability under the IC \cite{gauraha2017dual}. It is also be shown that features discarded by SAFE rules are the inactive constraints over the optimal dual solution \cite{wang2013lasso}, which justifies omitting them. KKT conditions imply:
\begin{equation} \lb{kkt}
\hat{\theta}^T X_i 
\begin{cases}
= \text{sign}(\hat{W}_i) \text{ if } \hat{W}_i \neq 0 \\
\in (-1, 1)  \text{ if } \hat{W}_i = 0
\end{cases}
\end{equation}
\Eref{kkt} gives us a direct way to compute the active dual set (which is the same as the active primal set with high probability) once we have $\hat{\theta}$. We discard the features for which $\hat{\theta}^T X_i \in (-1, 1)$ and retain the others. This does not give us a good fit of the solution, so to compute the coefficients accurately, we perform ridge regression ($\mathcal{P}(W) = \lambda_2 ||W||_2^2$ over these active features. This pseudocode for this is given in algorithm \ref{alg:dual}. We refer to this algorithm as `dual LASSO'.

\begin{algorithm}
	\caption{Sparse regression using dual LASSO} \label{alg:dual}
	\begin{algorithmic}
		\Require state parameters: $\bm{x} = x_i^t, \dot{\bm{x}} = \dot{x}_i^t$; LASSO penalty $\lambda$, ridge penalty $\lambda_2$
		\State Compute the primal LASSO solution: $\hat{W} = \min_{W} \left[ \( \dot{X} - XW \)^2 + \lambda ||W||_1 \right]$
		\State Compute the unique dual solution $\hat{\theta} = \dot{X} - X\hat{W} \lb{sta}$
		\State Compute dual active set (same as primal active with h.p.) $S^d = \{1 \leq j \leq m : \hat{\theta}^T X_j \notin (-1, 1) \}$
		\State Construct reduced feature matrix $\tilde{X}$ by only considering features whose indices are in $S^d$
		\State Solve ridge regression $W^* = \min_{W} \left[\( \dot{X} - XW \)^2 + \lambda_2 ||W||_2^2 \right]$
	\end{algorithmic}
\end{algorithm}
\vspace*{-0.1in}

\section{Sparse Regression Over Adaptive Feature Space} \label{sec:adap}

In this section, we look at cases where the feature library is not known. If we have no prior belief over the form of the equations, we may not be able to construct an efficient feature library. In such situations, learning this library from data might be the most advantageous choice.
%The most na\"ive choice is to blindly add new functions to the feature library until the some convergence metric is reached. This however tends to make the problem computationally very expensive and ill conditioned, as the higher order functions added (polynomials, for example) would be more and more correlated. This motivates us to move towards more principled and efficient approaches.
One approach is to make the use of orthogonal functions of some parametric family, such as polynomials to construct this library. This ensures no correlation between the features and affords maximum freedom to the algorithm. The drawback in this case is that the regressor may not be sparse over this feature library. % We propose the approach and analyze it in the following sections.

\subsection{Adaptive feature space growth through orthogonal regression} \label{sec:ortho_reg}

We propose the following approach for this purpose: Starting with an empty library, we recursively add a feature to it and compute the corresponding loss function of the resulting fit by using STRidge (\sref{stridge}). If the loss function decreases by more than a certain fraction, we keep this feature. Otherwise we discard it and look at the next orthogonal feature. Once every few of the addition timesteps, we perform a removal step.
That is, we discard the feature(s) that do not result in a significant increase in the loss function (if any).
%That is, we look at the existing feature library. We then try removing one-by-one each of the present features. If the loss function does not increase by a certain predetermined amount, we discard this feature.
This ensures that we do not keep lower order functions that may not be required to describe the equations as higher order functions are added.
%This also helps in the presence of noise, as noise may lead to incorrect lower order estimates, but should be corrected when considering higher order basis.
Our algorithm is inspired by previous greedy feature development algorithms by \citet{efron2004least} (LARS), \citet{zhang2009adaptive} (FoBa) etc.
%\citet{weisberg2005applied} (forward selection) and \citet{couvreur2000optimality} (backward elimination),
However these require pre-determined full possible feature space, whereas we construct new features on the fly. Once the equations are obtained in terms of these orthogonal polynomials, we distill their sparse forms by using symbolic equation simplification \cite{bailey2014automated}. 

\subsection{Sequentially thresholded ridge regression} \label{sec:stridge}

To compute regressors over the orthogonal feature space, we use sequentially thresholded ridge regression (STRidge), proposed by \citet{rudy2017data}. The idea is simple: we iteratively compute the ridge regression solution with decreasing penalty proportional to the condition number of $X$, and discard the components using scale based thresholding (\sref{fixed}).
% We iterate with ridge regression until there is no change in the feature space.
As the feature matrix is orthonormal by construction, the analytical solution is $W = \(1 + \lambda \)^{-1}X^T\dot{X}$.
%at every iteration of ridge regression, which is not expensive to compute. Note that $X^T\dot{X}$ only needs to be computed once. As the features are removed from $X$, we can remove the corresponding rows from $X^T\dot{X}$.
This algorithm is effective in choosing optimal features without necessitating sparsity.
The overall pseudocode for learning the governing equations through adaptively growing the feature library is given by algorithm \ref{alg:adap}, and the corresponding results are presented in \sref{res}.

\begin{algorithm}
	\caption{Learning the governing equations through adaptive growth of the feature library} \label{alg:adap}
	\begin{algorithmic}
		\Require state parameters: $\bm{x} = x_i^t, \dot{\bm{x}} = \dot{x}_i^t$; orthogonal family $F_j(\bullet)$; feature addition / removal thresholds: $r_{a} \ (\leq 1) , r_{r} \ (\geq 1), \lambda_0$; removal step frequency $k_r$
		\State Initialize: $X = \emptyset, W = \bm{0}, t = 0, \mathcal{L} = \infty$
		\While{True}
		\State $X_{t} = \text{append}(X, F_k(\bm{x}))$
		\State Solve the STRidge problem: $W_{t} = \text{STRidge}(\dot{X}, X_{t}, \lambda_0)$
		\State Compute the loss $\mathcal{L}_{t} = \( \dot{X} - X_{t} W_{t} \)^2$
		\If{$\mathcal{L}_t \leq r_a \mathcal{L}$}
		\State $X = X_t \ ; \ W = W_t$
		\EndIf
		\If{$\mod(k, k_r) == 0$}
		\For{$i = 1, \ldots, X.\text{shape}[1]$} \quad \quad \quad \quad \quad \quad \quad \quad \quad \ \ \ (number of columns of $X$)
		\State $X_t = \text{append}\(X[:, 1:i-1], X[:, i+1:\text{end}]\)$ \quad (ignore the $i^{th}$ column of $X$)
		\State Solve the STRidge problem: $W_{t} = \text{STRidge}(\dot{X}, X_{t}, \lambda_0)$
		\State Compute the loss $\mathcal{L}_{t} = \( \dot{X} - X_{t} W_{t} \)^2$
		\If{$\mathcal{L}_t \leq r_r \mathcal{L}$}
		\State $X = X_t \ ; \ W = W_t$
		\EndIf
		\EndFor
		\EndIf
		\State $k = k+1$.
		\State \textbf{break} if no change in feature space over multiple iterations.
		\EndWhile
		\State Perform symbolic simplification of $\dot{X} = XW$ to obtain the final form of the equations
	\end{algorithmic}
\end{algorithm}
%\vspace*{-0.1in}

\section{Results} \label{sec:res}

For the first part (\sref{fixed}), our testbed will be the Lorenz 63 system ($n = 3$), given by \eref{lorenz}.
% We first present the results for fixed feature library (\sref{fixed}), followed by those for the adaptive feature library growth (\sref{adap}).
%
\begin{equation}\lb{lorenz}
\dot{x} = 10(yz - x) \ ; \quad \quad \quad \dot{y} = x(28 - z) \ ; \quad \quad \quad \dot{z} = xy - 2.667 z
\end{equation}
For the first part, we consider three polynomial feature libraries with $p = 3, 10$ and $20$ ($m = 20, 286$ and $1771$). The idea behind considering larger orders ($p$) is that it highlights the poor performance of LASSO for highly correlated features.
%We tune the penalty parameter $\lambda$ by considering it to be $\approx \sqrt{T\log(m)}$ and the do a search.
\Fref{residual} shows the residual after the LASSO and the dual LASSO model fits. We can see that the residuals for both methods are comparable for all of the feature libraries, which empirically shows that the model fit even for a high degree of correlations is good for both the methods.
%\Fref{bar} tells the other side of the story, where we plot the number of non-zero (\ie selected) features in the equations for different $p$ values. We can see that LASSO has a much higher number of non-zero terms, and this number increases significantly with $p$ (and $m$), indicating instability of the solution.
\Fref{bar} plots the number of non-zero features in the equations for different $p$ values. LASSO has a much higher number of non-zero terms, and this number increases significantly with $p$ (and $m$), indicating instability of the solution.
Dual LASSO performs very well, and the number of present features does not change for the most part with $p$.
% It also turns out that the features present for the dual LASSO solution are exactly the ones in the original equation.
%\Fref{coeffs} plots the absolute weights for the components for the $p = 3$ case for the equation for $\dot{y}$.
\Fref{coeffs} plots the absolute weights for the components for the $p = 3$ case for the $\dot{y}$ equation.
Dual LASSO retrieves the correct features (with accurate weights), while LASSO detects the correct features but also detects high order features that have low weights and are highly correlated to each other. This serves as a great validation of the superiority of dual LASSO over conventional LASSO for model discovery.
Now we look at the results for \sref{adap}, on the quadratic Lorenz system (\eref{lorenz_quad}) \cite{eminauga2015modified}.
\begin{equation}\lb{lorenz_quad}
\dot{x} = 10(yz - x) \ ; \quad \quad \quad \dot{y} = x(28 - z) \ ; \quad \quad \quad \dot{z} = (xy)^2 - 2.667 z
\end{equation}
We start with an empty feature library and $W = \bm{0}$ and iteratively grow the feature space using algorithm \ref{alg:adap} using Legendre polynomials (denoted by $\leg_p(\bullet)$), $r_a = 0.75, r_r = 1.25, \lambda_0 = 1$ and removal step working every 10 addition steps ($k_r = 10$). The final obtained model is given in \eref{final_adap}. We use symbolic simplification \cite{bailey2014automated} to simply this model. Note that the model is not unique before symbolic simplification (as $\leg_0(x) = \leg_0(y) = \leg_0(z)$). We also do scale based thresholding (\sref{fixed}) once after the symbolic simplification to remove any terms that may remain due to approximate factorization. The results before and after scale based thresholding are given in \eref{fin_1}. We can see that this algorithm does a good job at identifying the correct governing equations by iteratively building the feature library, and does not include any incorrect features.
%
%\begin{eqnarray}\lb{final_adap}
%\dot{x} &=& 9.93 \leg_1(y) \leg_1(z) - 9.89 \leg_1(x) \nonumber \\
%\dot{y} &=& 27.66 \leg_1(x) - 1.04\leg_1(x)\leg_1(z) \\
%\dot{z} &=& 0.43 \leg_2(x) \leg_2(y) + 0.22 \leg_2(x) + 0.21 \leg_2(y) - 2.62 \leg_1(z) \nonumber \\ &&  + 2.09 \leg_0(x) - 0.22 \leg_0(y) - 1.95 \leg_0(z) \nonumber
%\end{eqnarray}
%\begin{eqnarray}\lb{final_adap}
%\dot{x} =& 9.93 \leg_1(y) \leg_1(z) - 9.89 \leg_1(x) \quad
%\dot{z} = &0.43 \leg_2(x) \leg_2(y) + 0.22 \leg_2(x) + 0.21 \leg_2(y) \\
%\dot{y} =& 27.66 \leg_1(x) - 1.04\leg_1(x)\leg_1(z) \quad \quad
%& - 2.62 \leg_1(z) + 2.09 \leg_0(x) - 0.22 \leg_0(y) - 1.95 \leg_0(z) \nonumber
%\end{eqnarray}

\vspace*{-0.1in}
\begin{align} \lb{final_adap}
\begin{aligned}
	\dot{x} =& 9.93 \leg_1(y) \leg_1(z) - 9.89 \leg_1(x) \\
	\dot{y} =& 27.66 \leg_1(x) - 1.04\leg_1(x)\leg_1(z)
\end{aligned}
\ \
\begin{aligned}
	\dot{z} = &0.43 \leg_2(x) \leg_2(y) + 0.22 \leg_2(x) + 0.21 \leg_2(y) \\
	 -&2.62 \leg_1(z) + 2.09 \leg_0(x) - 0.22 \leg_0(y) - 1.95 \leg_0(z)
\end{aligned}
\end{align}
%
%
%\begin{align} \lb{fin_1}
%\text{After symbolic simplification: }
%\begin{cases}
%\dot{x} = &9.93 yz - 9.89x \\
%\dot{y} = &27.66x - 1.04xz \\
%\dot{z} = &0.97(xy)^2 + 0.007(x^2 - y^2) \\
% &- 2.62z + 0.027
%\end{cases}
%\end{align}
%%
%\begin{align} \lb{fin_2}
%\text{After scale based thresholding: }
%\begin{cases}
%\dot{x} = 9.93 yz - 9.89x \\
%\dot{y} = 27.66x - 1.04xz \\
%\dot{z} = 0.9675(xy)^2 - 2.62z
%\end{cases}
%\end{align}
%
\begin{align} \lb{fin_1}
\underbrace{\begin{aligned}
\dot{x} &= 9.93 yz - 9.89x \\
\dot{y} &= 27.66x - 1.04xz \\
\dot{z} &= 0.97(xy)^2 + 0.007(x^2 - y^2) \\
&- 2.62z + 0.027
\end{aligned}}_{\text{after symbolic simplification}} \quad \implies \quad 
\underbrace{\begin{aligned}
\dot{x} &= 9.93 yz - 9.89x \\
\dot{y} &= 27.66x - 1.04xz \\
\dot{z} &= 0.9675(xy)^2 - 2.62z
\end{aligned}}_{\text{after scale based thresholding}}
\end{align}

\section{Conclusions and Future Work} \label{sec:concl}

In this work, we investigated LASSO, proposed dual LASSO and data driven feature learning approaches to solve the problem of discovering governing equations only from state parameter data.
% We first define the problem and the solution methodology, followed by the application and analysis of LASSO for this case. motivated to overcome the pitfalls of LASSO, we propose a new algorithm (called dual LASSO here) that relies on the uniqueness of the dual solution for the active set selection. Finally, we propose a new algorithm to compute the governing equations in the case when the feature library is not known. This algorithm relies on iteratively building the feature library using appropriate orthogonal functional basis and using sequentially thresholded ridge regression. We demonstrate the applications of the proposed theory on the classic Lorenz 63 system and also the quadratic Lorenz system.
Future work directions involve extending the ideas of feature library building where one can construct the function to be added through a mix of a larger family of orthogonal functions. Approaches to involve kernel compositionality \cite{duvenaud2013structure} to build these libraries can also be investigated. % Finally, it would be interesting to study the applications of these algorithms in the presence of model and observation noise, and to higher dimensional systems often encountered in physics and engineering.

\begin{figure}[thp!]
	\centering
	\vspace{0.135in}
	\subfloat[Residuals after the model fit\label{fig:residual}]{\includegraphics[trim=0 0 0 65, width=0.5\textwidth]{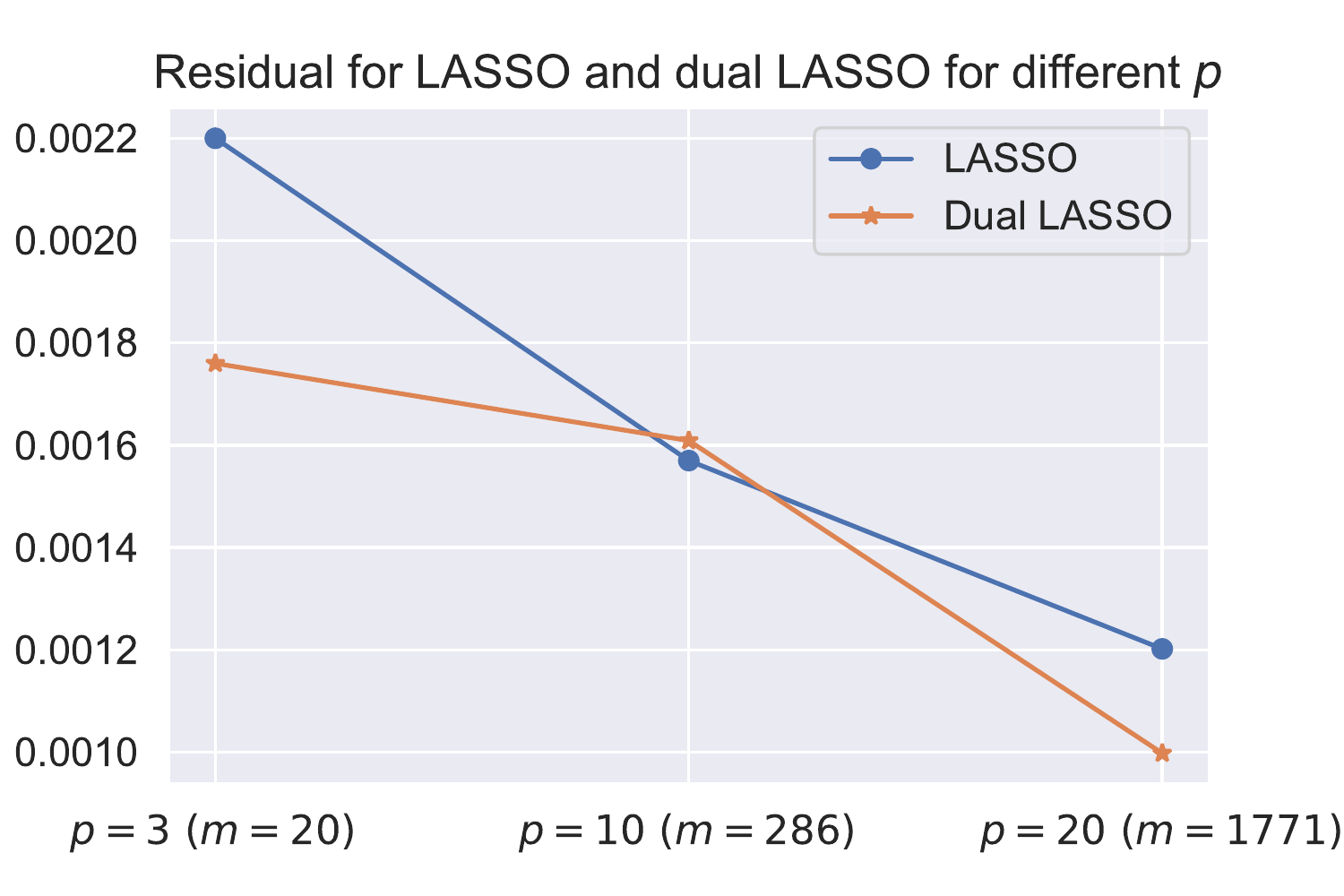}}
	\subfloat[Number of nonzero terms in the equations for $\dot{x}, \dot{y}, \dot{z}$\label{fig:bar}]{\includegraphics[trim=0 0 0 65,width=0.5\textwidth]{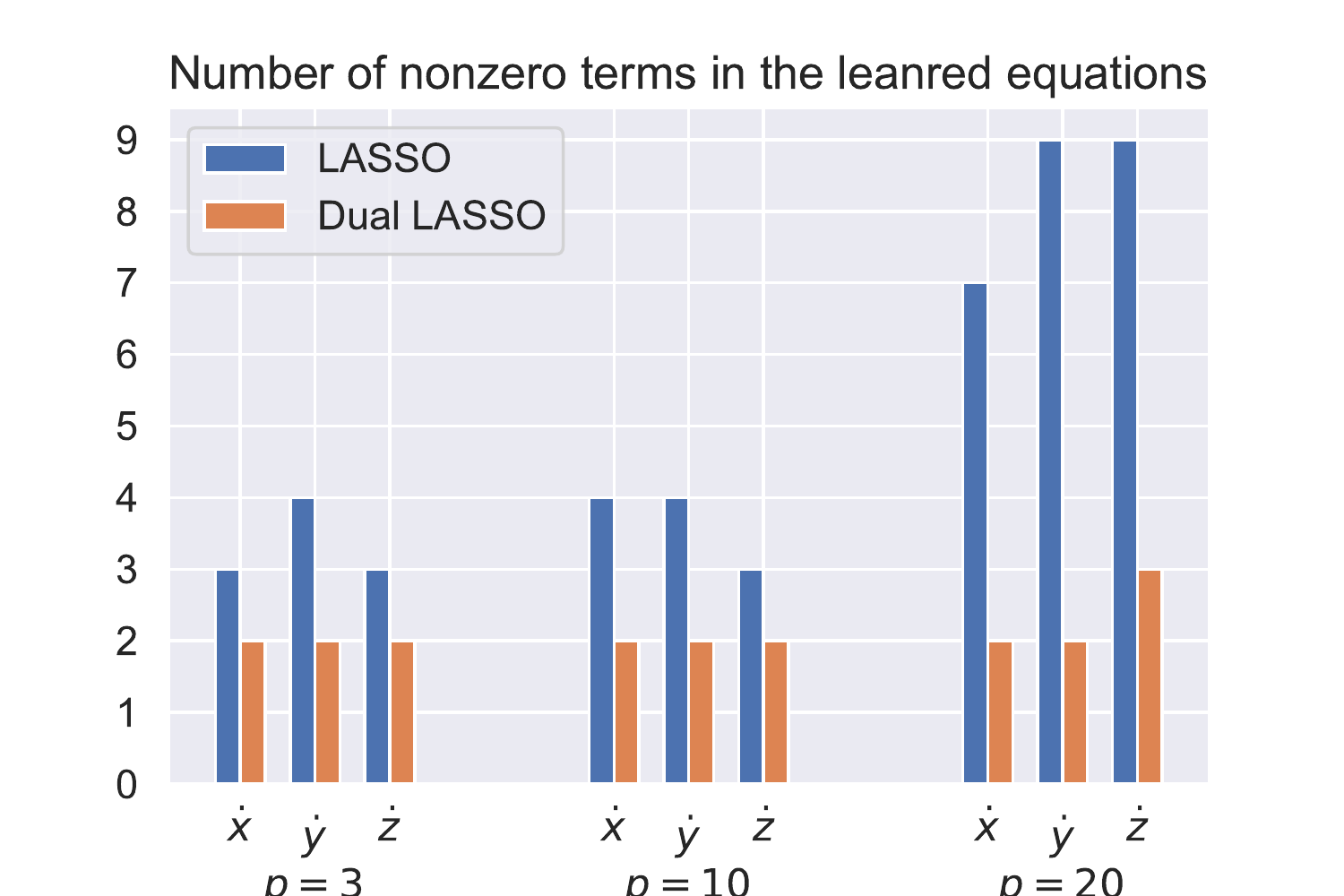}} \\
	\subfloat[The absolute weights in the equation for $\dot{y}$ \label{fig:coeffs}]{\includegraphics[trim=0 0 0 40,width=0.65\textwidth]{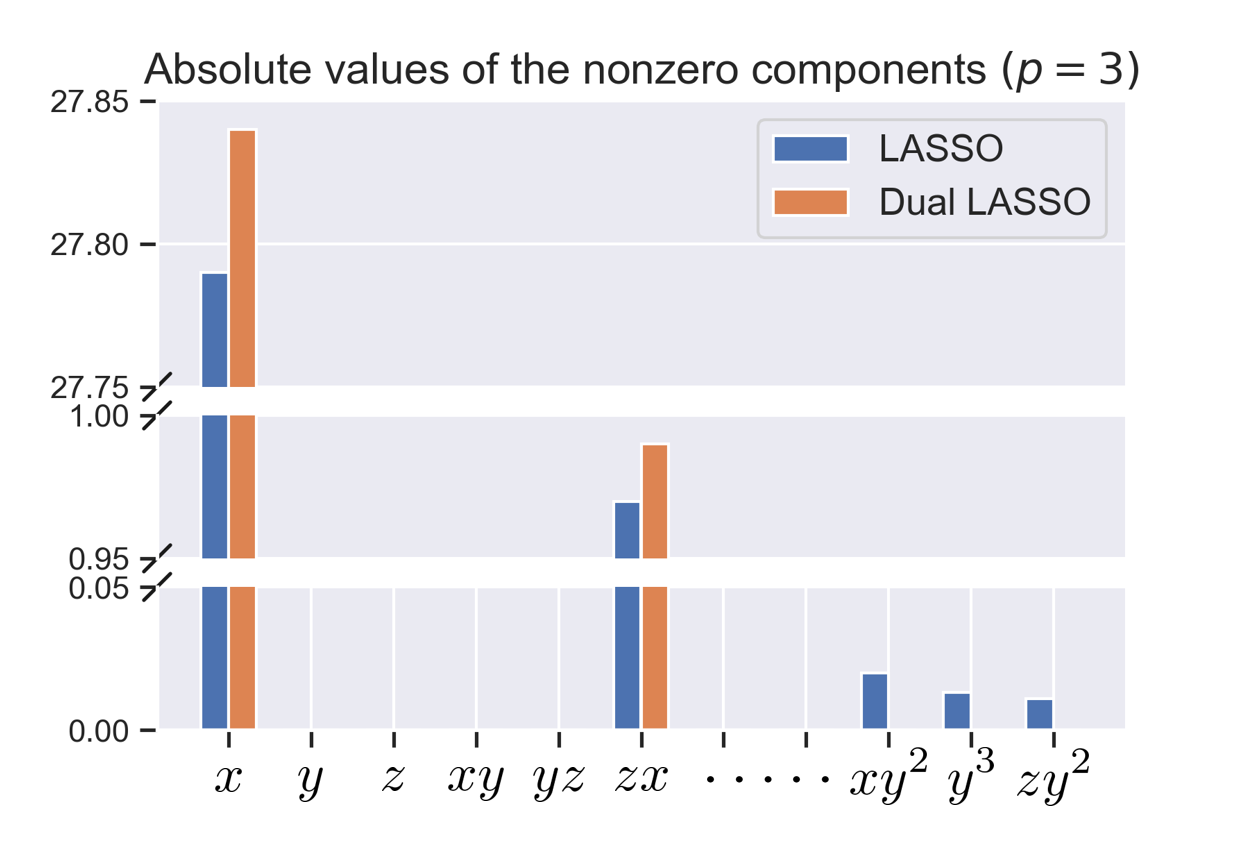}}
	\caption{Normalized (combined) residual after model fit, corresponding number of non-zero coefficients in the equation for $\dot{y}$, and the absolute weights for the components in the $\dot{y}$ eqation ($p = 3$).}
	\label{fig:Y_post}
\end{figure}

\bibliographystyle{plainnat}
\bibliography{biblio}

\end{document}